\documentclass[sigconf,nonacm]{acmart}
\usepackage{epsfig}
\usepackage{graphicx}
\usepackage{amsmath}

\usepackage{booktabs}
\usepackage{makecell}
\usepackage{multirow}
\usepackage{caption}
\usepackage{gensymb}
\usepackage[lined,boxed,commentsnumbered,ruled]{algorithm2e}
\usepackage{rotating}
\usepackage[caption=false]{subfig}
\usepackage{bigstrut} 
\usepackage{url}

\begin{document}

\title{End-to-End Video Object Detection with Spatial-Temporal Transformers}

\author{Lu He$^1$$^,$$^2$$^,$$^4$$^*$, Qianyu Zhou$^1$$^,$$^2$$^,$$^4$$^*$, Xiangtai Li$^3$$^,$$^4$$^*$, Li Niu$^1$,
Guangliang Cheng$^4$, \\
Xiao Li$^4$, Wenxuan Liu$^5$, Yunhai Tong$^3$,  Lizhuang Ma$^1$$^,$$^2$, Liqing Zhang$^1$\\
$^1$Department of Computer Science and Engineering, Shanghai Jiao Tong University, $^2$Qingyuan Research Institute, SJTU, $^3$Peking University, $^4$Sensetime Research,  $^5$UCLA \\
{\tt\small \{147258369, zhouqianyu, ustcnewly\}@sjtu.edu.cn}, 
\tt\small \{lxtpku,yhtong\}@pku.edu.cn
\\
\tt\small \{chengguangliang, lixiao\}senseauto.com, 
\tt\small 
\tt\small \{ma-lz, zhang-lq\}@cs.sjtu.edu.cn
}
\thanks{$^*$ The first three authors contribute equally to this work.}
\renewcommand{\shortauthors}{He, Zhou and Li, et al.}

\begin{abstract}

Recently, DETR and Deformable DETR have been proposed to eliminate the need for many hand-designed components in object detection while demonstrating good performance as previous complex hand-crafted detectors. However, their performance on Video Object Detection (VOD) has not been well explored. In this paper, we present TransVOD, an end-to-end video object detection model based on a spatial-temporal Transformer architecture. The goal of this paper is to streamline the pipeline of VOD, effectively removing the need for many hand-crafted components for feature aggregation, \emph{e.g.,} optical flow, recurrent neural networks, relation networks. Besides, benefited from the object query design in DETR, our method does not need complicated post-processing methods such as Seq-NMS or Tubelet rescoring, which keeps the pipeline simple and clean. In particular, we present temporal Transformer to aggregate both the spatial object queries and the feature memories of each frame. Our temporal Transformer consists of three components: Temporal Deformable Transformer Encoder (TDTE) to encode the multiple frame spatial details, Temporal Query Encoder (TQE) to fuse object queries, and Temporal Deformable Transformer Decoder to obtain current frame detection results. These designs boost the strong baseline deformable DETR by a significant margin (3\%-4\% mAP) on the ImageNet VID dataset. TransVOD yields comparable results performance on the benchmark of ImageNet VID. We hope our TransVOD can provide a new perspective for video object detection. Code will be made publicly available at \url{https://github.com/SJTU-LuHe/TransVOD}.
\end{abstract}





\maketitle

\section{Introduction}

Object detection is a fundamental task in computer vision and enables various applications in the real world, \emph{e.g.,} autonomous driving and robot navigation. Recently, deep convolution neural networks have achieved significant progress in object detection~\cite{ren15faster,ren2016faster,he17maskrcnn,lin2017feature,liu2016ssd,dai16rfcn}. Meanwhile, DETR~\cite{detr} and Deformable DETR~\cite{zhu2020deformable} remove many complex components such as NMS and dense anchors which makes the object detection a sparse set prediction problem. They also achieve competitive results with previous works. However, all these still-image detectors cannot be directly applied to video data, due to the appearance deterioration and changes of video frames, \emph{e.g.,} motion blur, video defocus, occlusion, rare poses. 

\begin{figure}[t]
    \centering
    \includegraphics[width=0.5\textwidth]{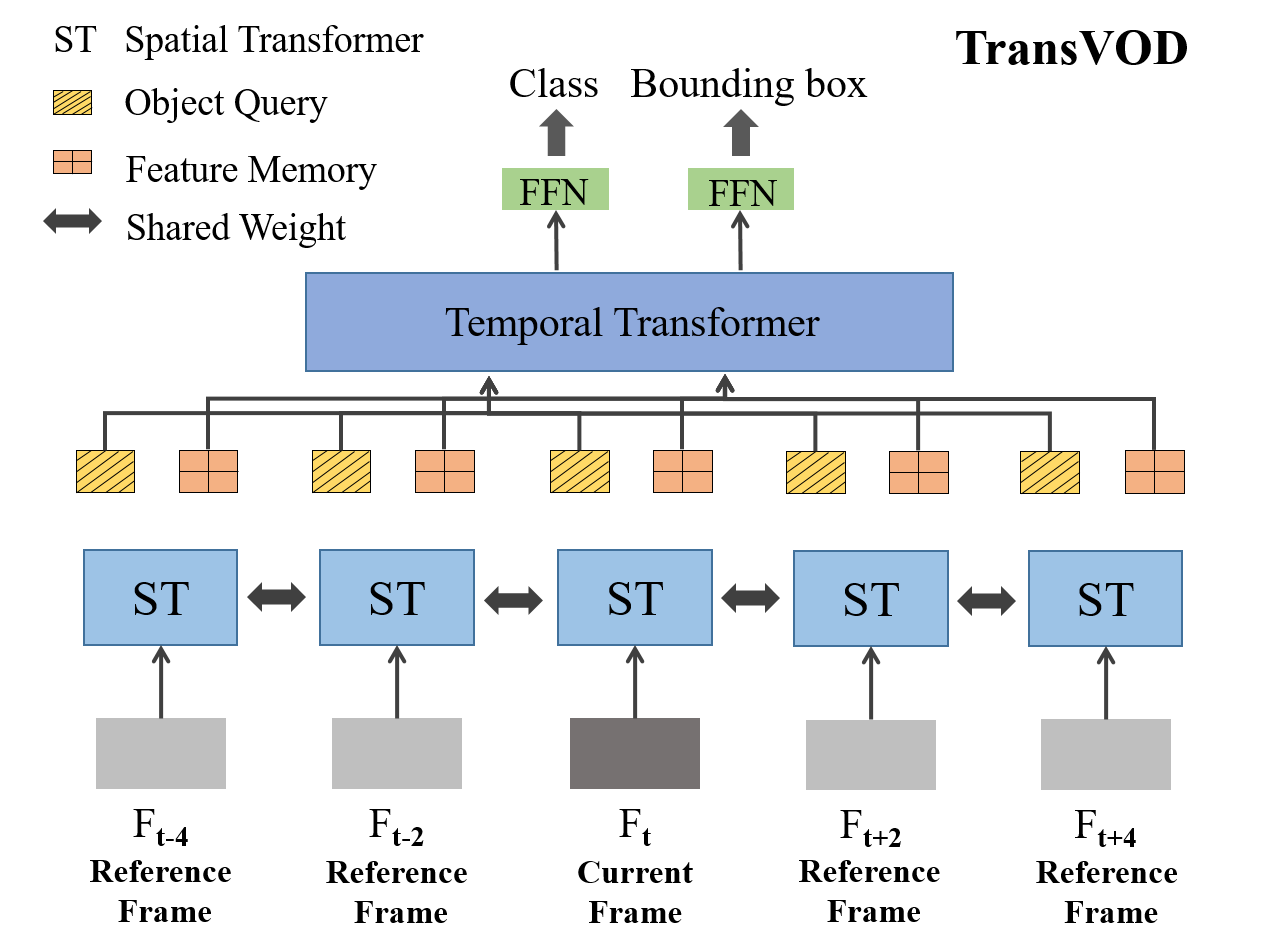} 
    \caption{\small Illustration of our proposed TransVOD. Our network is based on Spatial Transformer which outputs object query and feature memory of each frame. We propose a temporal Transformer to link both the spatial object queries and feature memories in a temporal dimension to obtain the results of the current frame. The final detection results are obtained via a shared feed forward network (FFN).}
    \label{fig:teaser}
\end{figure}

Thus, video object detection aims to detect all objects given a video clip. Previous video object detection approaches mainly leverage temporal information in two different manners. The first one relies on post-processing of temporal information to make the object detection results more coherent and stable ~\cite{han2016seq,kang2017t,belhassen2019improving,sabater2020robust,FocalLoss,tian2019fcos}. These methods usually apply a still-image detector to obtain detection results, then associate the results. Another line of approaches~\cite{yao2020video,jiang2020learning,han2020mining,han2020exploiting,lin2020dual,he2020temporal,chen2018optimizing,chen2020memory,sun2021mamba,guo2019progressive} exploits the feature aggregation of temporal information. Specifically, they mainly improve features of the current frame by aggregating that of adjacent frames or entire clips to boost the detection performance via specific operator design. In this way, the problems such as motion blur, part occlusion, and fast appearance change can be well solved. In particular, most of these methods~\cite{chen2018optimizing,chen2020memory,sun2021mamba,guo2019progressive} use two-stage detector Faster-RCNN~\cite{ren2016faster} or R-FCN~\cite{dai16rfcn} as the still-image baseline. 

Despite the gratifying success of these approaches, most of the two-stage pipelines for video object detection are over sophisticated, requiring many hand-crafted components, \emph{e.g.,} optical flow model~\cite{zhu17dff,zhu17fgfa,wang18manet,zhu18hp}, recurrent neural network~\cite{deng2019ogemn,chen2020memory,guo2019progressive}, deformable convolution fusion~\cite{bertasius18stsn,jiang2019video,he2020temporal}, relation networks~\cite{deng19rdn,chen2020memory,shvets19lltr}. In addition, most of them need complicated post-processing methods by linking the same object across the video to form tubelets and aggregating classification scores in the tubelets to achieve the state-of-the-art performance~\cite{han2016seq,kang2017t,belhassen2019improving,sabater2020robust}. Thus, it is in desperate need to build a \textit{simple yet effective} VOD framework in a fully end-to-end manner.

Transformers~\cite{wang2020end,detr,zhu2020deformable,dosovitskiy2020image,sun2020transtrack} have shown promising potential in the computer vision. Especially, DETR~\cite{detr,zhu2020deformable} simplifies the detection pipeline by modeling the object queries and achieves comparative performance with highly optimized CNN-based detectors. However, directly applying such a detector for video object detection also results in inferior results which will be shown in the experiment part. Thus, how to model the temporal information in a long-range video clip is a very critical problem. 

In this paper, our goal is to extend the DETR-like object detection into the video object detection domain. Our motivation has three aspects. Firstly, we observe that the video clip contains rich inherent temporal information, \emph{e.g.,} rich visual cues of motion patterns. Thus, it is natural to view video object detection as a sequence-to-sequence task with the advantages of Transformers~\cite{Vaswani17attention}. 
The whole video clip is like a sentence, and each frame contributes similarly to each word in natural language processing. Transformer can not only be used in inner each frame to model the interaction of each object but also can be used to link object along the temporal dimension. Secondly, object query is a one key component design in DETR~\cite{detr} which encodes instance-aware information. The learning process of DETR can be seen as the grouping process: grouping each object into an object query. Thus, these query embeddings can represent the instances of each frame and it is natural to link these sparse query embeddings via another transformer. Thirdly, the output memory from the DETR transformer encoder contains rich spatial information which can also be modeled jointly with query embeddings along the temporal dimension. 

Motivated by these facts, in this paper, we propose TransVOD, a new end-to-end video object detection model based on a spatial-temporal Transformer architecture. Our TransVOD views video object detection as an end-to-end sequence decoding/prediction problem.
For the current frame, as shown in Figure~\ref{fig:teaser}, it takes multiple frames as inputs and directly outputs current frame detection result via a Transformer-like architecture. In particular, we design a novel temporal Transformer to link each object query and memory encoding outputs simultaneously. Our proposed temporal Transformer mainly contains three components: Temporal Deformable Transformer Encoder (TDTE) to encode the multiple frame spatial details, Temporal Query Encoder (TQE) to fuse object query in one video clip, and Temporal Deformable Transformer Decoder (TDTD) to obtain final detection results of the current frame. The TDTE efficiently aggregates spatial information via Temporal Deformable Attention and avoids background noise. The TQE first adopts a coarse-to-fine strategy to select relevant object queries in one clip and fuse such selected queries via several self-attention layers~\cite{Vaswani17attention}. TDTD is another decoder that takes the outputs of TDTE and TQE, and outputs the final detection results. These modules are shared for each frame and can be trained in an end-to-end manner. Then temporal information can be well explored via such design. We carry out extensive experiments on ImageNet VID dataset~\cite{russakovsky2015imagenet}. Compared with the single-frame baseline~\cite{zhu2020deformable}, our TransVOD achieves significant improvements (3-4\% mAP). We also present detailed ablation studies and provide more visual analysis of our proposed methods. Our main contributions are summarized as following:

\begin{itemize}
    \item We propose TransVOD, a novel Transformer-based framework for end-to-end video object detection. Unlike the previous hand-crafted methods, our TransVOD treats the VOD as a sequence encoding and decoding task. The framework is significantly different from existing methods, simplifying the overall pipeline of VOD considerably.   
    
    \item Based on the original DETR design, we extend the DETR-like object detector via a temporal Transformer. Our proposed temporal Transformer contains three main components including Deformable Transformer Encoder (TDTE), Temporal Query Encoder (TQE), and Temporal Deformable Transformer Decoder (TDTD). The former two modules efficiently link the spatial information and object queries respectively while the latter one outputs the current detection result. 
    
    \item We carry out experiments on ImageNet VID datasets to verify the effectiveness of our TransVOD. Our proposed temporal Transformer improves the strong still image detector by a large margin (3\%-4\% mAP). We achieve comparable results with previous state-of-the-art methods without any post-processing.
    TransVOD achieves 79.9\% mAP on ImageNet VID validation set using ResNet50 as backbone and 81.9\% mAP using ResNet101 as backbone.
    Moreover, we perform extensive ablation studies and visual analysis on each component.

\end{itemize}

\section{Related work}

\noindent
\textbf{Video Object Detection:} The VOD task requires detecting objects in each frame and linking the same objects across frames. State-of-the-art methods typically develop sophisticated pipelines to tackle it. One common solution~\cite{yao2020video,jiang2020learning,han2020mining,han2020exploiting,lin2020dual,he2020temporal,chen2018optimizing,chen2020memory,sun2021mamba,guo2019progressive} to amend this problem is feature aggregation that enhances per-frame features by aggregating the features of nearby frames. Earlier works adopt flow-based warping to achieve feature aggregation. Specifically, DFF~\cite{zhu17fgfa}, FGFA~\cite{zhu17fgfa} and THP~\cite{zhu18hp} all utilize the optic flow from FlowNet~\cite{dosovitskiy2015flownet} to model motion relation via different temporal feature aggregation. To calibrate the pixel-level features with inaccurate flow estimation, MANet~\cite{wang18manet} dynamically combines pixel-level and instance-level calibration according to the motion in a unified framework. Nevertheless, these flow-warping-based methods have several disadvantages: 1) Training a model for flow extraction requires large amounts of flow data, which may be difficult and costly to obtain. 2) integrating a flow network and a detection network into a single model may be challenging due to multi-task learning. Another line of attention-based approaches utilize self-attention~\cite{vaswani2017attention} and non-local~\cite{wang2018non} to capture long-range dependencies of temporal contexts. SELSA~\cite{wu19selsa} treats video as a bag of unordered frames and proposes to aggregate features in the full-sequence level. STSN~\cite{bertasius18stsn} and TCENet~\cite{he2020temporal} propose to utilize deformable convolution to aggregate the temporal contexts within a complicated framework with so many heuristic designs. RDN~\cite{deng19rdn} introduce a new design to capture the interactions across the objects in spatial-temporal context. LWDN~\cite{jiang2019video} adopts a memory mechanism to propagate and update the memory feature from keyframes to keyframes. OGEMN~\cite{deng2019ogemn} present to use object-guided external memory to store the pixel and instance-level features for further global aggregation. MEGA~\cite{chen2020memory} considers aggregating both the global information and local information from the video and presents a long-range memory. Despite the great success of these approaches, most of the pipelines for video object detection are too sophisticated, requiring many hand-crafted components, \emph{e.g.,} optic flow model, memory mechanism, or recurrent neural network. In addition, most of them need complicated post-processing methods such as Seq-NMS~\cite{han2016seq}, Tubelet rescoring~\cite{kang2017t}, Seq-Bbox Matching~\cite{belhassen2019improving}  or REPP~\cite{sabater2020robust} by linking the same object across the video to form tubelets and aggregating classification scores in the tubelets to achieve the state-of-the-art. Instead, we aim to build a \textit{simple and end-to-end trainable} VOD framework without these designs.

\noindent
\textbf{Transformers:} Recently, Transformers~\cite{detr,zhu2020deformable, wang2020end,dosovitskiy2020image,sun2020transtrack,meinhardt2021trackformer} have raised great attention in the computer vision. DETR~\cite{detr} builds a fully end-to-end object detection system based on Transformers, which largely simplifies the traditional detection pipeline. It also achieves on par performances compared with highly-optimized CNN-based detectors~\cite{ren2016faster}. However, it suffers from slow convergence and limited feature spatial resolution, Deformable DETR~\cite{zhu2020deformable} improves DETR by designing a deformable attention module, which attends to a small set of sampling locations as a pre-filter for prominent key elements out of all the feature map pixels. Our work is inspired by DETR~\cite{detr} and Deformable DETR~\cite{zhu2020deformable}. The above works show the effectiveness of Transformers in image object detection tasks. There are several con-current works that applied Transformer into video understanding, \emph{e.g.,} Video Instance Segmentation (VIS)~\cite{vis_dataset}, multi-object tracking (MOT). TransTrack~\cite{sun2020transtrack} introduces a query-key mechanism into the multi-object tracking model while Tranformer~\cite{meinhardt2021trackformer} directly adds track query for MOT. However, both only leverage limited temporal information, \emph{i.e.,} just the previous frame. We suppose that this way can not fully use enough temporal contexts from a video clip. VisTR~\cite{wang2020end} views the VIS task as a direct end-to-end parallel sequence prediction problem. The targets of a clip are disrupted in such an instance sequence and directly performing target assignment is not optimal. Instead, we aim to link the outputs of the spatial Transformer, \emph{i.e.,} object query, through a temporal Transformer, which acts in a completely different way from VisTR~\cite{wang2020end}. To our knowledge, there are no prior applications of Transformers to video object detection (VOD) tasks so far. It is intuitive to see that the Transformers’ advantage of modeling long-range dependencies in learning temporal contexts across multiple frames for VOD task. Here, we propose the TransVOD, leveraging both the spatial Transformer and the temporal Transformer and then providing an affirmative answer to that.

\section{Method}


\noindent
\textbf{Overview.} We will first review the related work including both DETR~\cite{detr} and Deformable DETR~\cite{zhu2020deformable}. Then we will give detailed descriptions of our proposed Temporal Transformer which contains three key components: Temporal Deformable Transformer Encoder (TDTE), Temporal Query Encoder (TQE), and Temporal Transformer Decoder (TTD). Finally, we describe the network architecture details. 

\begin{figure*}[h]
    \centering
    \includegraphics[width=0.85\textwidth]{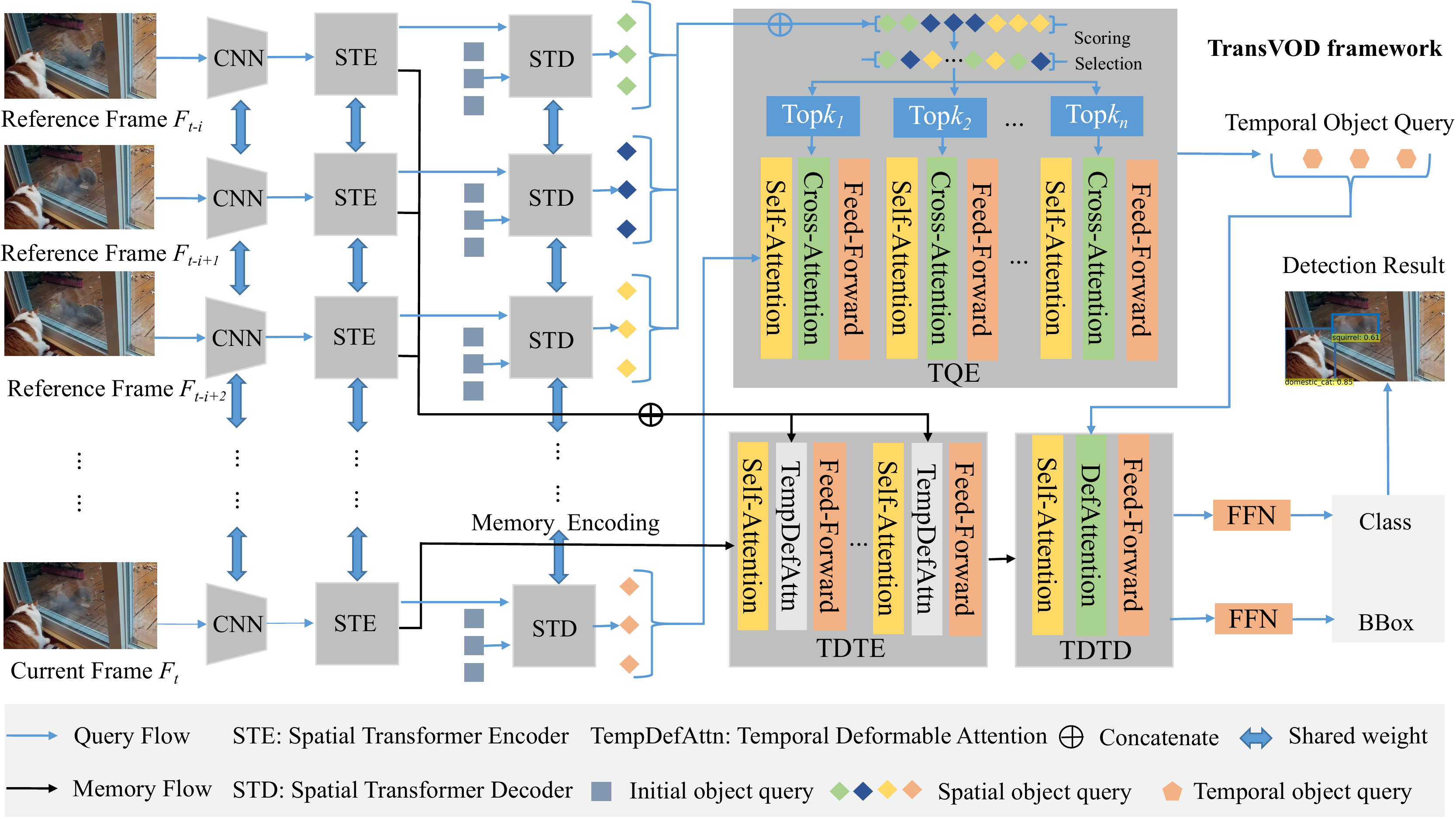} 
    \caption{\small \textbf{The whole pipeline of TransVOD.} A shared CNN backbone extracts feature representation of multiple frames. Next, a series of shared Spatial Transformer Encoder (STE) produces the feature memories and these memories are linked and fed into Temporal Deformable Transformer Encoder (TDTE). Meanwhile, the Spatial Transformer Decoder (STD) decodes the spatial object queries. Naturally, we use a Temporal Query Encoder (TQE) to model the relations of different queries and aggregate these output queries, thus we can enhance the object query of the current frame. Both the temporal query and the temporal memories are fed into the Temporal Deformable Transformer Decoder (TDTD) to learn the temporal contexts across different frames. }
    \label{fig:framework}
\end{figure*}

\subsection{Revisiting DETR and Deformable DETR}
DETR~\cite{detr} formulates object detection as a set prediction problem. A CNN backbone~\cite{he2016deep} extracts visual feature maps $f  \in \mathbb{R}^{C \times H \times W}$ from an image $I \in \mathbb{R}^{3 \times H_0 \times W_0}$, where $H,W$ are the height/width of the input image and $H_0,W_0$ are the height and width of the visual feature map, respectively. The visual features augmented with position embedding $f_{pe}$ would be fed into the encoder of the Transformer. Self-attention would be applied to $f_{pe}$ to generate the key, query, and value features $K, Q, V$ to exchange information between features at all spatial positions. To increase the feature diversity, such features would be split into multiple groups along the channel dimension for the multi-head self-attention. Let $q\in\Omega_q$ indexes a query element with representation feature $z_q \in R^{C}$, and $k\in\Omega_k$ indexes a key element with representation feature $x_k \in R^{C}$, where $C$ is the feature dimension, $\Omega_q$ and $\Omega_k$ specify the set of query and key elements, respectively. Then the multi-head attention feature is calculated by
\begin{align}
    \text{MultiHeadAttn}(z_q, x) = \sum_{m=1}^{M} W_m \big[\sum_{k\in\Omega_k} A_{mqk} \cdot W'_m x_k \big],
    \label{eq:co-attention} 
\end{align}

where $m$ indexes the attention head, $W'_m \in R^{C_v \times C}$ and $W_m \in R^{C \times C_v}$ are of learnable weights ($C_v = C/M$ by default). The attention weights $A_{mqk}$ are normalized as:
\begin{align}
   A_{mqk} \propto \exp\{\frac{z_q^T U_m^T~ V_m x_k}{\sqrt{C_v}}\}, \sum_{k\in\Omega_k} A_{mqk} = 1,
\end{align}
in which $U_m, V_m \in R^{C_v \times C}$ are also learnable weights. To disambiguate different spatial positions, the representation features $z_q$ and $x_k$ are usually of the concatenation/summation of element contents and positional embeddings. 
The decoder's output features of each object query is then further transformed by a Feed-Forward Network (FFN) to output class score and box location for each object. Given box and class prediction, the Hungarian algorithm is applied between predictions and ground-truth box annotations to identify the learning targets of each object query for one to one matching. 

Following DETR, Deformable DETR~\cite{zhu2020deformable} replaces the multi-head self-attention layer with a deformable attention layer to efficiently sample local pixels rather than all pixels. Moreover, to handle missing small objects, they also propose a cross attention module that incorporates multi-scale feature representation. Due to the fast convergence and computation efficiency, we adopt Deformable DETR as our still image Transformer detector baseline.


\subsection{TransVOD Architecture} The overall TransVOD architecture is shown in Figure~\ref{fig:framework}. It takes multiple frames in a video clip as the input and outputs the detection results for the current frame. It contains four main components: Spatial Transformer for single frame object detection to extract both object queries and compact features representation (memory for each frame), Temporal Deformable Transformer Encoder (TDTE) to fuse memory outputs from Spatial Transformer, Temporal Query Encoder (TQE) to link objects in each frame along the temporal dimension and Temporal Deformable Transformer Decoder (TDTD) to obtain final output for the current frame.

\noindent
\textbf{Spatial Transformer:} We choose the recently proposed Deformable DETR~\cite{zhu2020deformable} as our still image detector. In particular, to simplify complex designs in Deformable DETR, we \emph{do not} use multi-scale feature representations in both encoders and decoders. We only use the last stage of the backbone as the input of the deformable Transformer. The modified detector includes Spatial Transformer Encoder and Spatial Transformer Decoder which encodes each frame $F_{t}$ (including Reference Frame and Current Frame) into two compact representations: Spatial Object Query $Q_{t}$ and Memory Encoding $E_{t}$ (shown in blue arrow and black arrow in Figure~\ref{fig:framework}). 

\noindent
\textbf{Temporal Deformable Transformer Encoder:}
The goal of the Temporal Deformable Transformer Encoder is to encode the spatial-temporal feature representations and provide the location cues for the final decoder output. Since most adjacent features contain similar appearance information, using naive Transformer encoder~\cite{detr,Vaswani17attention} directly may bring much extra computation (most useless computation on object background). Deformable attention~\cite{zhu2020deformable} samples only partial information efficiently according to the learned offset field. Thus we can link these memory encodings $E_{t}$ through the deformable attention in a temporal dimension. 

The core idea of the temporal deformable attention modules is that we only attend to a small set of key sampling points around a reference, and thus we can aggregate the features across difference frames more efficiently. The multi-head deformable attention is as follows:
\begin{align}
    \text{TempDefAttn}(z_q, \hat{p}_q, \{x^l\}_{l=1}^{L}) &=\sum_{m=1}^{M} W_m \big[\sum_{l=1}^{L} \sum_{k=1}^{K} A_{mlqk} \nonumber \\
    & x^l(\phi_l(\hat{p_q}) + \Delta p_{mlqk} )\big], \label{eq:def-attention}
\end{align}
where, $m$ indexes the attention head, $l$ indexes the frame sampled from the same video clip, and $k$ indexes the sampling point, and $\Delta p_{mlqk}$ and $A_{mlqk}$ denote the sampling offset and attention weight of the $k^\text{th}$ sampling point in the $l^\text{th}$ frame and the $m^\text{th}$ attention head, respectively.
The scalar attention weight $A_{mlqk}$ lies in $[0, 1]$, normalized by $\sum_{l=1}^{L} \sum_{k=1}^{K} A_{mlqk} = 1$. $\Delta p_{lmqk} \in R^2$ are of 2-d real numbers with unconstrained range. As $p_q + \Delta p_{mlqk}$ is
fractional, bilinear interpolation is applied as in \cite{dai2017deformable} in computing $x(p_q + \Delta p_{mlqk})$. 
Both $\Delta p_{mlqk}$ and $A_{mlqk}$ are obtained via linear projection over the query feature $z_q$. Here, we use normalized coordinates $\hat{p}_q \in [0, 1]^2$ for the clarity of scale formulation, in which the normalized coordinates $(0, 0)$ and $(1, 1)$ indicate the top-left and the bottom-right image corners, respectively. Function $\phi_{l}(\hat{p}_q)$  re-scales the normalized coordinates $\hat{p}_q$ to the input feature map of  $l$-th frame. The multi-frame temporal deformable attention samples $LK$ points from $L$ feature maps instead of $K$ points from single-frame feature maps. Supposing that there exist a total $M$ attention heads in each Temporal Transformer encoder layer. Following \cite{zhu2020deformable}, the query feature $z_q$ of each frame is fed to a linear projection operator of $3MK$ channels, where the first $2MK$ channels encode the sampling offsets $\Delta p_{iqk}$, and the remaining $MK$ channels are fed to a $\operatorname{Softmax}$ operator to obtain the attention weights $A_{iqk}$.

\noindent
\textbf{Temporal Query Encoder:}
As mentioned in the previous part, learnable object queries can be regarded as a kind of non-geometric deep anchor, which automatically learns the statistical features of the whole still image datasets during the training process. It means that the spatial object queries are not related to temporal contexts across different frames. Thus, we propose a \emph{simple yet effective} encoder to measure the interactions between the objects in the current frame and the objects in reference images. 

Our key idea is to link these spatial object queries in each frame via a temporal Transformer, and thus learn the temporal contexts across different frames. We name our module Temporal Query Encoder (TQE). TQE takes all the spatial object queries from reference frames to enhance the spatial output query of the current frame, and it outputs the Temporal Query for the current frame. Moreover, inspired by previous works~\cite{deng19rdn}, we design a coarse-to-fine spatial object query aggregation strategy to progressively schedule the interactions between the current object query and the reference object queries. The benefit of such a coarse-to-fine design is that we can reduce the computation cost to some extent. 

We combine the spatial object query from all reference frames, denoted as $Q_{ref}$. Then, we perform the scoring and selection in a coarse-to-fine manner. Specifically, we use an extra Feed Forward Network (FFN) to predict the class logits and after that, we calculate the sigmoid value of that: $p=Sigmoid [ FFN(Q_{ref}) ]$. Then, we sort all the reference points by is $p$ value and select the top-confident $k$ values from these reference points. Considering that the shallow layer may learn more detailed information, while there is less information in the deep layers, we perform a coarse-to-fine selection. In other words, the shallow layers should select more confident queries, and the last layers should choose less credible object queries. The selected values are fed to feature refiners to interact with the object queries extracted from different frames, calculating the co-attention with the output of the current frame. The decoder layers with cross-attention modules play
the role of a cascade feature refiner which updates output queries of each spatial Transformer iteratively. The refined temporal object query is the input of the Temporal Deformable Transformer Decoder (TDTD).

\noindent
\textbf{Temporal Deformable Transformer Decoder}
This decoder aims to obtain the current frame output according to both outputs from TDTE (fused memory encodings) and TQE (temporal object queries). Given the aggregated feature memories $\hat{E}$  and the temporal queries $\hat{O_q}$, our Temporal Deformable Transformer Decoder (TDTD) performs co-attention between online queries and the temporal aggregated features.
The deformable co-attention~\cite{zhu2020deformable} of the temporal decoder layer is shown as follows:
\begin{equation}
\text{DeformAttn}(z_q, p_q, x) = \sum_{m=1}^{M} W_m \big[\sum_{k=1}^{K} A_{mqk} \cdot W'_m x(p_q + \Delta  p_{mqk})\big],
\label{eq:single_deform_attn_fun}
\end{equation}
where $m$ indexes the attention head, $k$ indexes the sampled keys, and $K$ is the total sampled key number ($K \ll HW$).
$p_{mqk}$ and $A_{mqk}$ denote the sampling offset and attention weight of the $k^\text{th}$ sampling point in the $m^\text{th}$ attention head, respectively.
The scalar attention weight $A_{mqk}$ lies in the range of $[0, 1]$, normalized by $\sum_{k=1}^{K} A_{mqk} = 1$. $\Delta p_{mqk} \in R^2$ are of 2-d real numbers with unconstrained range. As $p_q + \Delta p_{mqk}$ is
fractional, bilinear interpolation is applied as in \citet{dai2017deformable} in computing $x(p_q + \Delta p_{mqk})$. 
Both $\Delta p_{mqk}$ and $A_{mqk}$ are obtained via linear projection over the query feature $z_q$. In our implementation, the query feature $z_q$ is fed to a linear projection operator of $3MK$ channels, where the first $2MK$ channels encode the sampling. The output of TDTD is sent to one feed-forward network for the final classification and box regression as the detection results of the current frame.


\noindent{\bf Loss Function:} Original DETR~\cite{detr} avoid post-processing and adopt a one-to-one label assignment rule. Following~\cite{stewart2016end, detr, zhu2020deformable}, we match predictions from STD/TDTD with ground truth by Hungarian algorithm~\cite{kuhn1955hungarian} and thus the entire training process of spatial Transformer is the same as original DETR. 
The temporal Transformer uses similar loss functions given the box and class prediction output by two FFNs. 
The matching cost is defined as the loss function. Following \cite{detr, zhu2020deformable, sun2020sparse}, the loss function is as follows:.
\begin{align}
    \mathcal{L} = \lambda_{cls}\cdot\mathcal{L}_{\mathit{cls}}+\lambda_{L1} \cdot \mathcal{L}_{\mathit{L1}}+\lambda_{giou}\cdot\mathcal{L}_{\mathit{giou}} 
\end{align}
$\mathcal{L}_{\mathit{cls}}$ represents focal loss~\cite{FocalLoss} for classification. $\mathcal{L}_{\mathit{L1}}$ and $\mathcal{L}_{\mathit{giou}}$ represent L1 loss and generalized IoU loss~\cite{GIoU} in for localization. $\lambda_{cls}$, $\lambda_{L1}$ and $\lambda_{giou}$ are coefficients of them. We balance these loss functions following the same setting in ~\cite{zhu2020deformable}.

\section{Experiment}


\begin{table*}[htbp]
\footnotesize
\begin{center}
\begin{tabular}{c|c|c|c|c|c}
\toprule
Methods & Venue & Backbone & Base Detector & Aggregate frames  & mAP(\%) \\
\midrule
DFF~\cite{zhu17dff} & CVPR'17 & ResNet-50 & R-FCN & 10 &70.4\\
FGFA~\cite{zhu17fgfa} & ICCV'17 & ResNet-50 & R-FCN & 21  &74.0\\
\midrule
Single Frame Baseline~\cite{ren2016faster} & NeurIPS'15 & ResNet-50 & Faster-RCNN & 1 & 71.8 \\
RDN~\cite{deng19rdn}  & ICCV'19 & ResNet-50 & Faster-RCNN & 3  & 76.7\\
MEGA~\cite{chen2020memory} & CVPR'20 & ResNet-50 & Faster-RCNN & 9  & 77.3\\
\midrule
Single Frame Baseline~\cite{zhu2020deformable} & ICLR'21 & ResNet-50 & Deformable DETR & 1 & 76.0 \\
TransVOD & - & ResNet-50 & Deformable DETR & 9 & 79.0 \\
TransVOD & - & ResNet-50 & Deformable DETR & 15 & \textbf{79.9} \\
\bottomrule
\end{tabular}
\end{center}
\caption{Comparison with the state-of-the-art on ImageNet VID (ResNet 50 backbone).}
\label{table:mainresult_r50}
\end{table*}%
\begin{table*}[htbp]
\footnotesize
\begin{center}
\begin{tabular}{c|c|c|c|c|c}
\toprule
Methods & Venue & Backbone & Base Detector & Params (M) & mAP(\%) \\
\midrule
Single Frame Baseline~\cite{dai2016r} & NeurIPS'16 & ResNet-101 & R-FCN & 59.6 & 73.6 \\ 
DFF~\cite{zhu17dff} & CVPR'17 & ResNet-101 & R-FCN &96.6 &73.0\\                
D$\&$T~\cite{feichtenhofer17dt} & ICCV'17 & ResNet-101 & R-FCN &100.4  &75.8 \\  
FGFA~\cite{zhu17fgfa} & ICCV'17 & ResNet-101 & R-FCN &-  &76.3\\             
MANet~\cite{wang18manet} & ECCV'18  & ResNet-101 & R-FCN &-  &78.1\\         
THP~\cite{zhu18hp} & CVPR'18  & ResNet-101+DCN & R-FCN &-  &78.6\\           
STSN~\cite{bertasius18stsn} & ECCV'18  & ResNet-101+DCN & R-FCN &-  &78.9\\   
PSLA~\cite{guo2019progressive} & ICCV'19 & ResNet-101+DCN & R-FCN &72.2  & 80.0 \\
OGEMN~\cite{deng2019ogemn} & ICCV'19  & ResNet-101+DCN & R-FCN &-  & 80.0 \\ 
\multirow{2}{*}{LSTS~\cite{jiang2020learning}} & ECCV'20 & ResNet-101 & R-FCN &64.5  &  77.2 \\
 & ECCV'20 & ResNet-101+DCN & R-FCN & 65.5  & 80.2 \\
\midrule
Single Frame Baseline~\cite{ren2016faster} & NeurIPS'15 & ResNet-101 & Faster RCNN &- & 76.7 \\
ST-Lattice~\cite{chen2018optimizing} & CVPR'18  & ResNet-101 &  Faster RCNN &>100 &79.0\\
STCA~\cite{luo2019object} & Arxiv'19  & ResNet-101 &  Faster RCNN &-  & 80.3 \\
SELSA~\cite{wu19selsa} & ICCV'19   & ResNet-101 & Faster RCNN &-  & 80.3\\
LRTR~\cite{shvets19lltr} & ICCV'19 & ResNet-101 & Faster RCNN  &-  & 81.0 \\
RDN~\cite{deng19rdn}  & ICCV'19 & ResNet-101 & Faster RCNN &-  & 81.8\\
MEGA~\cite{chen2020memory} & CVPR'20 & ResNet-101 & Faster RCNN &-  & \textbf{82.9}\\
\midrule
Single Frame Baseline~\cite{zhou2019objects} & Arxiv'19 & ResNet-101 & CenterNet &-  &73.6\\ %
CHP~\cite{xu2020centernet} & ECCV'20 & ResNet-101 & CenterNet &-  &76.7\\ %
\midrule
Single Frame Baseline~\cite{zhu2020deformable} & ICLR'21 & ResNet-101 &  Deformable DETR & 53.4 & 78.3 \\
TransVOD & - & ResNet-101  & Deformable DETR & 58.9 & \textbf{81.9} \\
\bottomrule
\end{tabular}
\end{center}
\caption{Comparison with the state-of-the-art on ImageNet VID (ResNet 101 backbone).}
\label{table:mainresult_r101}
\end{table*}%

\noindent
\textbf{Overview} In this section, we first introduce the experimental setup for the video object detection task, including the dataset and evaluation protocols. Then, we present the implementation details of the proposed TransVOD. We also compare our proposed method with several other state-of-the-art video object detection methods. Finally, we perform several ablation studies and provide both visualization results and analysis on the ImageNet VID~\cite{russakovsky2015imagenet} validation set to verify the effectiveness of the proposed network. 

\subsection{Experimental Setup}

\noindent\textbf{Datasets.} We empirically conduct experiments on the ImageNet VID dataset~\cite{russakovsky2015imagenet} which is a large-scale benchmark for video object detection. It contains 3862 training videos and 555 validation videos
with annotated bounding boxes of 30 classes. Since the ground truth of the official
testing set is not publicly available, we follow the widely adopted setting in previous work~\cite{zhu17fgfa, wang18manet, deng19rdn, wu19selsa} where we train our models using a combination of ImageNet VID and DET datasets~\cite{russakovsky2015imagenet} and measure the performance on the validation set using the mean average precision (mAP) metric. 

\noindent \textbf{Implementation detail.} 
We use ResNet-50~\cite{he2016resnet} and ResNet-101~\cite{he2016resnet} as the network backbone. Following original Deformable DETR~\cite{zhu2020deformable}, the optimizer is AdamW~\cite{loshchilov2017decoupled} with batch size 2, initial Transformer’s learning rate $2\times 10^{-4}$, the backbone’s $2\times 10^{-5}$, and weight decay $10^{-4}$. All Transformer weights are initialized with Xavier init~\cite{glorot2010understanding}, and the backbone ImageNet-pretrained~\cite{deng2009imagenet} model with frozen batch-norm layers~\cite{ioffe2015batch}. The number of initial object query is set as 300. We use the pre-trained model of our modified Deformable DETR on COCO dataset~\cite{COCO_dataset}. Following previous work~\cite{chen2020memory}, we use the same data augmentation including random horizontal flip, random resizing the input images such that the shortest side is at least 600 while the longest at most 1000. We train the network for 10 (\emph{resp.}, 12) epochs and the learning rate drops at the 8-th (\emph{resp.}, 10-th)  epochs when using the ResNet-50 (\emph{resp.}, ResNet-101) as the network backbone. In the inference phase, we do not need any sophisticated post-processing method, which largely simplifies the pipeline of the VOD. 



\subsection{Main Results}
Table~\ref{table:mainresult_r50} and Table~\ref{table:mainresult_r101} show the comparison results from existing state-of-the-art VOD methods with different backbones (\emph{e.g.}, ResNet 50 and ResNet 101), respectively. Note that for fair comparison,
the comparison is performed under the same circumstance. To be noted, we only show the state-of-the-art methods \textbf{without} any post-processing in Table. \ref{table:mainresult_r50} and Table~\ref{table:mainresult_r101}.
\begin{figure*}[htb]
    \centering
    \includegraphics[width=0.85\textwidth]{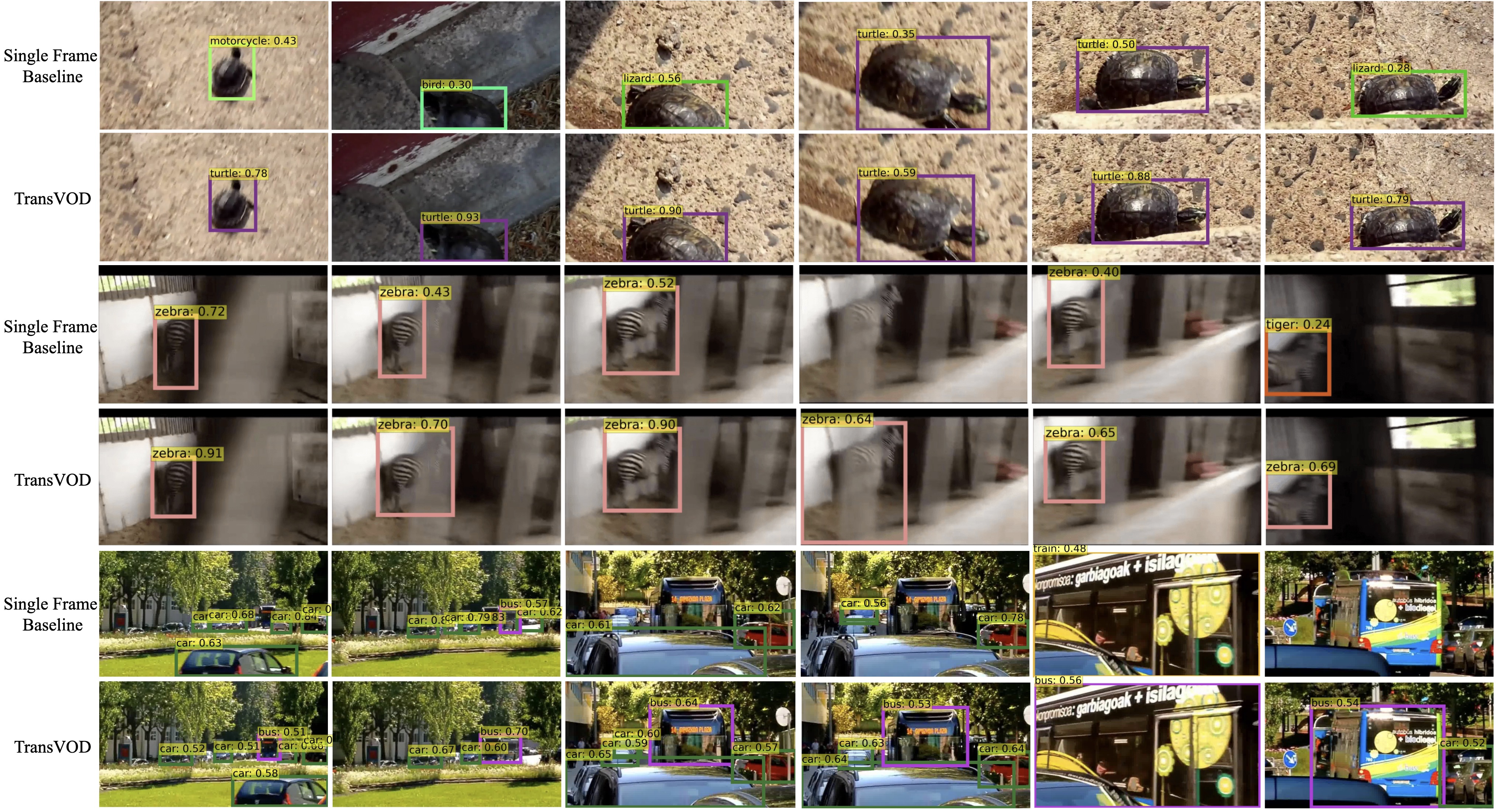} 
    \caption{\small The visual results of single frame baseline method~\cite{zhu2020deformable} and TransVOD in different scenarios. Compared with single frame baseline, our proposed TransVOD shows better and consistent detection results in the cases of rare pose(top two rows), motion blur(middle two rows), and part occlusion(last two rows), respectively.}
    \label{visual prediction results}
\end{figure*}
As shown in Table~\ref{table:mainresult_r50}, the results under the same backbone ResNet-50 demonstrate that our proposed TransVOD achieves the best performance against the state-of-the-art methods by a large margin. In particular, the mAP can achieve 79.9 $\%$ with ResNet-50, which makes 2.6 $\%$ absolute improvement over the best competitor MEGA~\cite{chen2020memory}. As shown in Table~\ref{table:mainresult_r101}, when equipped with a stronger backbone ResNet-101, the mAP of our TransVOD is further boosted up to 81.9\%, which outperforms most  state-of-the-art methods~\cite{zhu17dff,feichtenhofer17dt,zhu17fgfa,wang18manet,zhu18hp}. Specifically, our model is remarkably better than FGFA~\cite{zhu17fgfa} (76.3$\%$ mAP) and MANet~\cite{wang18manet} (78.1$\%$ mAP), which both aggregate features based on optical flow estimation, and the mAP improvements are +5.6$\%$ mAP and +3.8$\%$ mAP respectively.
Moreover, our proposed method boosts the strong baseline \emph{i.e.,} deformable DETR~\cite{zhu2020deformable} by a significant margin (\textbf{3$\%$ $\sim$ 4$\%$ mAP}). Furthermore, the parameter count (58.9M) is fewer than other video object detectors using an optical flow network (\emph{e.g.,} around 100M), which also indicates that our method is more friendly for mobile devices.




\subsection{Visualization Analysis}
\noindent \textbf{Visual detection results:} As shown in Figure ~\ref{visual prediction  results}, we show the visual detection results of still image detector Deformable DETR~\cite{zhu2020deformable} and our proposed TransVOD in odd and even rows, respectively. The still image detector is easy to cause false detection (\emph{e.g.,} turtle detected as a lizard) and missed detection (\emph{e.g.,} zebra not detected), in the case of motion blur, part occlusion. Compared with still image detectors, our method can effectively model the long-range dependencies across different video frames to enhance the features of the detected image. Thus, our proposed TransVOD can not only increase the confidence of correct prediction but also effectively reduce the number of cases that are missed or falsely detected.

\begin{figure}[h]
    \includegraphics[width=0.45\textwidth]{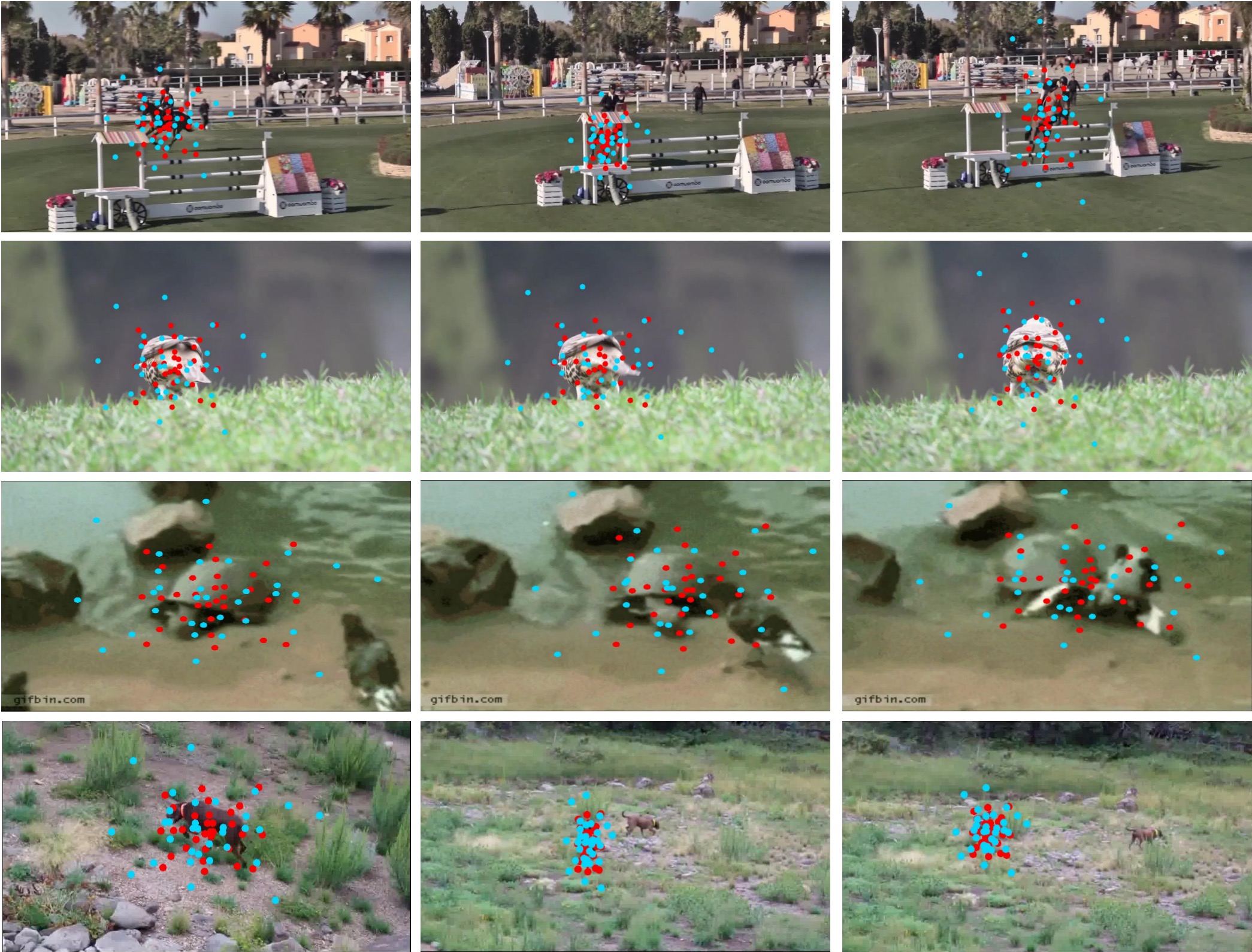} 
    \caption{\small The visualization of the deformable cross-attention in the last spatial Transformer decoder layer and temporal Transformer decoder layer. We visualize the sampling locations of the temporal object query and corresponding spatial object query in one picture. Each sampling point of the temporal object query is marked as a red-filled circle, while the blue circle represents the sampling point of the spatial object query. }
    \label{fig:sample_vis_res}
\end{figure}

\noindent \textbf{Visual sampling locations of object query:} To further explore the advantages of TQE, we visualize the sampling locations of both spatial object query and temporal object query in Figure~\ref{fig:sample_vis_res}. The sample locations indicate the most relevant context for each detection. As shown in that figure, for each frame in each clip, our temporal object query \textbf{has more concentrated and precise results} on fore-ground objects while the original spatial object query has more diffuse results. This proves that our temporal object query is more suitable for detecting objects in video. 

\begin{table}[t]
\caption{Ablation on effectiveness of each component in our TransVOD with ResNet 50 backbone.}
\label{table:ablation_camix}
\begin{center}
\begin{tabular}{cccc|cc} \toprule
Single Frame Baseline & TDTE & TQE & TDTD & mAP \\
\midrule
$\surd$  &  &  & & 76.0 \\
$\surd$  & $\surd$ & & $\surd$ & 77.1 \\
$\surd$  &  & $\surd$ &  & 78.9 \\
$\surd$  &   & $\surd$ & $\surd$ & 79.3 \\
$\surd$  & $\surd$  & $\surd$ & $\surd$  & 79.9 \\
\bottomrule
\end{tabular}
\end{center}
\label{table:ablation_component}
\end{table}

\subsection{Ablation Study}

To demonstrate the effect of key components in our proposed method, we conduct extensive experiments to study how they contribute to the final performance using ResNet-50 as backbone.  \\
\noindent \textbf{Effectiveness of each component:}
Table~\ref{table:ablation_component} summarizes the effects of different design components on ImageNet VID dataset. Temporal Query Encoder (TQE), Temporal Deformable Transformer Encoder (TDTE), and Temporal Deformable Transformer Decoder (TDTD) are three key components of our proposed method. The single-frame baseline Deformable DETR ~\cite{zhu2020deformable} is 76.0$\%$. By only adding TQE, we boost the mAP with an additional +2.9 $\%$, which demonstrates that TQE can effectively measure the interaction among the objects in different video frames. And then, by adding the TDTD and TDTE sequentially, we boost the mAP with an additional +0.4$\%$ and +0.6$\%$, achieving 79.3$\%$ and 79.9$\%$, respectively. These improvements show the effects of individual components of our proposed approach.
\begin{table}[htbp]
\footnotesize
\begin{center}
\begin{tabular}{c|c|c|c|c|c|c}
\toprule
Number of encoder layers in TQE & 1 & 2 & 3  & 4 & 5 & 6  \\
\midrule
mAP(\%)  & 78.8  & 79.4  & 79.6 & 79.6  & 79.7 &  79.7 \\

\bottomrule
\end{tabular}
\end{center}
\caption{Ablation on the number of encoder layers in TQE.}
\label{table:tqea_query_layer}
\end{table}%
\begin{table}[htbp]
\footnotesize
\begin{center}
\begin{tabular}{c|c|c|c|c|c}
\toprule
Number of encoder layers in TDTE & 0 & 1 & 2  & 3 & 4\\
\midrule
mAP(\%) & 77.0 & 77.7 & 77.6 & 77.8 & 77.7 \\

\bottomrule
\end{tabular}
\end{center}
\caption{Ablation on the number of encoder layers in TDTE}
\label{table:tdte_encoder_layer}
\end{table}%
\begin{table}[htbp]
\footnotesize
\begin{center}
\begin{tabular}{c|c|c|c|c|c|c|c}
\toprule
Number of layers in TDTD & 0 & 1 & 2  & 3 & 4 & 5 & 6\\
\midrule
mAP(\%) & 77.8 & 78.2 & 77.7 & 77.1 & 76.2 & 74.8 & 72.3 \\

\bottomrule
\end{tabular}
\end{center}
\caption{Ablation on the number of layers in TDTD.}
\label{table:tdtd_decoder_layer}
\end{table}%

\noindent 
\textbf{Number of encoder layers in TQE:}
Table~\ref{table:tqea_query_layer} shows the ablation study on the number of encoder layers in TQE. It shows that the best result occurs when the number query layer is set to 5. When the number of layers is up to 3, the performance is basically unchanged. Thus, we use 3 encoder layers in our final method.



\noindent \textbf{Number of encoder layers in TDTE:}
Table~\ref{table:tdte_encoder_layer} illustrates the ablation study on the number of encoder layers in TDTE. The highest mAP on the ImageNet VID validation dataset is achieved when the number of TDTE encoder layers is set to 1, which means that more TDTE encoder layers bring no benefits to the final performance. This experiment also proves the claim that aggregating the feature memories in a temporal dimension via deformable attention is useful for learning the temporal contexts across different frames.

\noindent \textbf{Number of decoder layers in TDTD:}
Table ~\ref{table:tdtd_decoder_layer} illustrates the ablation study on the number of decoder layers in TDTD. The basic setting is 4 reference frames, 1 encoder layer in TQE, and 1 encoder layer in TDTE. The results indicate that only one decoder layer in TDTD is needed.
\begin{table}[htbp]
\footnotesize
\begin{center}
\begin{tabular}{c|c|c|c|c|c|c}
\toprule
k & 25 & 50 & 100  & 200 & 300\\
\midrule
mAP(\%) & 78.0 & 78.1 & 78.3 & 77.9 &  77.7 \\
\bottomrule
\end{tabular}
\end{center}
\caption{Ablation of top k spatial object query in TQE with only one decoder layer.}
\label{table:_top_k}
\end{table}%
\begin{table}[htbp]
\footnotesize
\begin{center}
\begin{tabular}{c|c|c|c|c|c|c|c|c|c|c}
\toprule
k1& 30 &30 &30 & 50  &50 & 80 & 80 & 100 & 100 & 100 \\
k2& 20 &20 &30 & 30  &50 & 50 & 80 & 50 & 80 & 100 \\
k3& 10 &20 &30 & 20  &50 & 20 & 80 & 20 & 30  & 100 \\
\midrule
mAP(\%) & 79.7 & 79.6 & 79.3 & 79.6 &  79.5 & 79.9 & 79.7 &79.5 & 79.8 & 79.1\\
\bottomrule
\end{tabular}
\end{center}
\caption{Ablation of top k spatial object query numbers with three encoder layers. Our coarse-to-fine strategy has better results.}
\label{table:k3}
\end{table}%
\begin{table}[htbp]
\footnotesize
\begin{center}
\begin{tabular}{c|c|c|c|c|c|c}
\toprule
Number of reference frames & 2 & 4 & 8  & 10 & 12 & 14\\
\midrule
mAP(\%) & 77.7 & 78.3  & 79.0  & 79.1 & 79.0& 79.3 \\
\bottomrule
\end{tabular}
\end{center}
\caption{Ablation on number of  reference frames.}
\label{table:num_reference_image}
\end{table}%

 \\
\noindent \textbf{Number of top $k$ object queries in TQE:}
To verify the effectiveness of our coarse-to-fine Temporal Query Aggregation strategy, we conduct ablation experiments in Table~\ref{table:_top_k} and Table~\ref{table:k3} to study how they contribute to the final performance. All the experiments in each table are conducted under the same setting. The first experiment is that when we use 1 encoder layer in TQE with 4 reference frames, the best performance is achieved when we choose the top 100 spatial object queries for each reference frame. The second experiment is conducted in a multiple TQE enocder layers case, \emph{i.e.,} 3 encoder layers in TQE. We denote the Fine-to-fine (F2F) selection by using a small number of spatial object queries in each TQE encoder layer. coarse-to-coarse (C2C)  means selecting a large number of spatial object queries when performing the aggregation in each layer. Our proposed coarse-to-fine aggregation strategy is using a larger number of spatial object queries in the shallow layers and a smaller number of spatial object queries in the deep layers to conduct the query aggregation. The results in Table~\ref{table:k3} show that our coarse-to-fine aggregation strategy is superior to both the coarse-to-coarse selection and fine-to-fine selection. 

\noindent \textbf{Number of reference frames:} The Table~\ref{table:num_reference_image} illustrates the ablation study on number of reference. The basic setting is 3 encoder layers in TQE, 1 encoder layer in TDTE, and 1 decoder layer in TDTD. As shown in Table~\ref{table:num_reference_image}, the mAP improves when the number of reference frames increases, and it tends to stabilize when the number is up to 8.

\section{Conclusion}
In this paper, we proposed a novel video object detection framework, namely TransVOD, which provides a new perspective of feature aggregation by leveraging spatial-temporal Transformers. TransVOD effectively removes the need for many hand-crafted components and complicated post-processing methods. Our core idea is to aggregate both the spatial object queries and the memory encodings in each frame via Temporal Transformers. In particular, our temporal Transformer consists of three components: Temporal Deformable Transformer Encoder (TDTE) to efficiently encode the multiple frame spatial details, Temporal Query Encoder (TQE) to fuse spatial object queries obtaining temporal object queries, and Temporal Deformable Transformer Decoder (TDTD) to obtain current frame detection results. These designs boost the strong baseline deformable DETR by a significant margin (3\%-4\% mAP) on ImageNet VID dataset. Extensive experiments are conducted to study and verify the effectiveness of core components. Without bells and whistles, we achieve comparable performance on the benchmark of the ImageNet VID. 
To our knowledge, our work is the first one that applies the Transformer to video object detection tasks. We hope that similar approaches can be applied to more video understanding tasks in the future. 

\bibliographystyle{ACM-Reference-Format}
\bibliography{egbib}



\end{document}